# Predicting brain age with deep learning from raw imaging data results in a reliable and heritable biomarker

James H Cole[1], Rudra PK Poudel[2], Dimosthenis Tsagkrasoulis[3], Matthan WA Caan[4], Claire Steves[5], Tim D Spector[5], Giovanni Montana[2,3*]

**Author affiliations:**

[1]Computational, Cognitive & Clinical Neuroimaging Laboratory, Division of Brain Sciences, Imperial College London, London, UK.

[2]Department of Biomedical Engineering, King's College London, London, UK.

[3]Department of Mathematics, Imperial College London, London, UK.

[4]Department of Radiology, Academic Medical Center, Amsterdam.

[5]Department of Twin Research & Genetic Epidemiology, King's College London, London, UK.

**Corresponding author:** Giovanni Montana. Postal address: Department of Biomedical Engineering, King's College London. St Thomas' Hospital, The Rayne Institute 3rd Floor, Lambeth Wing St Thomas' Hospital, London SE1 7EH. E-mail: giovanni.montana@kcl.ac.uk

**Abstract**

Machine learning analysis of neuroimaging data can accurately predict chronological age in healthy people and deviations from healthy brain ageing have been associated with cognitive impairment and disease. Here we sought to further establish the credentials of 'brain-predicted age' as a biomarker of individual differences in the brain ageing process, using a predictive modelling approach based on deep learning, and specifically convolutional neural networks (CNN), and applied to both pre-processed and raw T1-weighted MRI data.

Firstly, we aimed to demonstrate the accuracy of CNN brain-predicted age using a large dataset of healthy adults (N = 2001). Next, we sought to establish the heritability of brain-predicted age using a sample of monozygotic and dizygotic female twins (N = 62). Thirdly, we examined the test-retest and multi-centre reliability of brain-predicted age using two samples (within-scanner N = 20; between-scanner N = 11). CNN brain-predicted ages were generated and compared to a Gaussian Process Regression (GPR) approach, on all datasets. Input data were grey matter (GM) or white matter (WM) volumetric maps generated by Statistical Parametric Mapping (SPM) or raw data.

CNN accurately predicted chronological age using GM (correlation between brain-predicted age and chronological age r = 0.96, mean absolute error [MAE] = 4.16 years) and raw (r = 0.94, MAE = 4.65 years) data. This was comparable to GPR brain-predicted age using GM data (r = 0.95, MAE = 4.66 years). Brain-predicted age was a significantly heritable phenotype for all models and input data ($h^2$ = 0.50-0.84). Brain-predicted age showed high test-retest reliability (intraclass correlation coefficient [ICC] = 0.90-0.98). Multi-centre reliability was more variable within high ICCs for GM (0.83-0.96) and poor-moderate levels for WM and raw data (0.51-0.77).

Brain-predicted age represents an accurate, highly reliable and genetically-valid phenotype, that has potential to be used as a biomarker of brain ageing. Moreover, age predictions can be accurately generated on raw T1-MRI data, substantially reducing computation time for novel data, bringing the process closer to giving real-time information on brain health in clinical settings.

# 1. Introduction

The human brain changes across the adult lifespan. This process of *brain ageing* occurs in accord with a general decline in cognitive performance, *cognitive ageing* . Although the changes associated with brain ageing are not explicitly pathological, with increasing age comes increasing risk of neurodegenerative disease and dementia (Abbott, 2011). However, the wide range of onset ages for age-associated brain diseases indicates that the effects of ageing on the brain vary greatly between individuals. Thus, advancing our understanding of brain ageing and identifying biomarkers of the process are vital to help improve detection of early-stage neurodegeneration and predict age-related cognitive decline.

One promising approach to identifying individual differences in brain ageing derives from the research showing that neuroimaging data, most commonly T1-weighted MRI, can be used to accurately predict chronological age in healthy individuals, using machine learning . By 'learning' the correspondence between patterns in structural or functional neuroimaging data and an age 'label', machine-learning algorithms can formulate massively high-dimensional regression models, fitting large neuroimaging datasets as independent variables to predict chronological age as the dependent variable. The resulting brain-based age predictions are generally highly accurate, particularly when algorithms learn from large training datasets and are applied to novel or 'left-out' data (i.e., test datasets).

Neuroimaging-derived age predictions have been explored in the context of different brain diseases. By training models on healthy individuals, brain-based predictions of age can then be made in independent clinical samples. If 'brain-predicted age' is greater than an individual's chronological age, this is thought to reflect some aberrant accumulation of age-related changes to the brain. The degree of this 'added' brain ageing can be simply quantified by subtracting chronological age from brain-predicted age. This approach is being used more frequently and has demonstrated increased brain-predicted age in adults with mild cognitive impairment who progress to Alzheimer's , after traumatic brain injury (Cole et al., 2015), in schizophrenia  and diabetes (Franke et al., 2013). At the same, brain-predicted age has been used to demonstrate protective influences on brain ageing, including meditation (Luders et al., 2016) and increased levels of education and physical exercise (Steffener et al., 2016). Evidently, the extent to which one's brain resembles the typical structure or function appropriate for one's age can be affected by both positive and negative influences. By conceptualising brain ageing in this manner, highly-complex multivariate datasets and statistical procedures can be reformulated into an intuitively straightforward and widely-applicable biomarker. However, the practicality of using such a marker clinically, its reliability and relevance for normal variation in brain ageing need to further demonstrated.

One hindrance to clinical applications for neuroimaging generally is the time needed for image 'post-processing' after acquisition (referred to as 'pre-processing' by neuroimagers), which can take hours or days, while clinical decisions often need to occur in minutes or less. Regardless of learning algorithm, previous brain-predicted age studies have required several pre-processing stages. Such steps are typically a sequence of data transformations that produce a representation of the original images that is sufficiently structured, compact and informative to support machine learning. These include the removal of non-brain tissue (i.e., skull stripping or brain extraction), affine or non-linear image registration, interpolation and smoothing. While pre-processing may reduce noise and permit voxelwise inter-individual statistical comparisons, there are numerous additional assumptions required for any pre-processing pipeline. These assumptions are often not met, particularly when analysing brain images containing gross pathology and can even be an increased source of error. Recently, however, modelling methods that require little or no image pre-processing have become available, so-called 'deep learning'.

The resurgence of interest in artificial neural networks for learning data representations, deep learning, offers a new way of approaching statistical modelling in neuroimaging, thanks to improvements in computing infrastructure. When sufficiently large volumes of data are available, no 'hand-engineering' (i.e., manually selecting *a priori* which features should be used as input) is needed as the deep learning algorithm is able to infer a compact representation of the data, starting only with raw images as input, that is optimally tailored for the particular predictive modelling task at hand. In this respect, deep learning offers several practical advantages for high-dimensional prediction tasks, that should enable the learning of both physiologically-relevant representations and latent relationships (Plis et al., 2014). Of particular interest to us is the potential for deep learning techniques, such as convolutional neural networks (CNN), to make predictions from raw, unprocessed neuroimaging data, thus obviating the reliance on time-consuming pre-processing and improving the clinical applicability of models of brain ageing.

Beyond improving clinical applicability, a biomarker of brain ageing needs to relate to naturally occurring variation, such as that caused by genetic factors. Many aspects of brain ageing and susceptibility to age-related brain disease are thought to be under genetic influence . Therefore, demonstrating a brain ageing biomarker is sensitive to genetic influences gives some external, genetic, validity to the measure. Furthermore, if a neuroimaging biomarker is significantly heritable, this motivates further research into specific candidate genes, or sets of genes, that may affect this aspect of brain ageing. These candidate genes can then, in turn, provide biological targets for pharmacological interventions which aim to improve brain health in older adults.

Another important facet of any biomarker is reliability. If a biomarker is to be evaluated longitudinally, in clinical trials or research settings, to track change over time, establishing test-retest reliability is vital. Furthermore, as many neuroimaging studies are now international collaborative efforts, data collection often takes place across multiple scanning sites. Therefore, between-scanner reliability, which indicates that a method of obtaining a biomarker is generalizable to data acquired from other sites, is of increasing importance.

In this work, we sought to establish the credentials of CNN-predicted age as a potential biomarker of brain ageing in three different ways: 1) Demonstrate that CNNs can accurately predict age using structural neuroimaging data and compare predictions using pre-processed and 'raw' input data; 2) Establish the heritability of brain-predicted age using a sample of monozygotic and dizygotic twins; 3) Assess both the test-retest (i.e. within-scanner) and multi-centre (i.e. between-scanner) reliability of brain-predicted age.

## 2. Materials and Methods

### 2.1. Datasets

All neuroimaging data used in the study were T1-weighted MRI scans. Details of the participants in the specific samples and the respective acquisition parameters used are outlined below:

#### 2.1.1. Brain-predicted age evaluation cohort

The evaluation of the accuracy of age modelling using neuroimaging was conducted using the Brain-Age Normative Control (BANC) dataset. This cohort consisted of N = 2001 healthy individuals (male/female = 1016/985, mean age = 36.95± 18.12, age range 18-90 years). These data were compiled from 14 publicly-available sources (see Supplementary material), made available via various data-sharing initiatives. All participants were screened to be free from major neurological or psychiatric diagnoses, according to local study protocols. All data were acquired at either 1.5T or 3T using standard T1-weighted sequences (full details in supplementary material). Each contributing study was ethically approved, as was subsequent data-sharing. Informed consent was obtained at each local study site in accordance with local guidelines.

#### 2.1.2. Heritability assessment sample

Participants for heritability assessment were individuals from the UK Adult Twin Registry, who were invited to take part in a neuroimaging sub-study. A total of 62 female individuals were scanned (mean age = 61.86 ± 8.36), including 27 monozygotic twin pairs and 4 dizygotic twin pairs. All participants were free from major neurological or psychiatric diagnoses and contraindications to MRI scanning. A Philips Achieva 3T was used to acquired T1-weighted 3D turbo field echo (TFE) MRI with the following parameters: TE = 3.21 ms, TR = 6.89 ms, flip angle = 8°, field-of-view = 240 mm, 133 slices of 1.2 mm thickness, in-plane resolution = 1.07 x 1.07 mm. Each participant provided written and informed consent for academic use of the data. Experiments were approved by the National Research Ethics Service (NRES) Committee London - Westminster.

#### 2.1.3. Within-scanner reliability sample

A total of 20 participants (male/female = 12/8, mean age at first scan = 34.05 ± 8.71) took part in the STudy Of Reliability of MRI (STORM) at Imperial College London. Participants were scanned an average of 28.35 ± 1.09 days apart. All participants were free from major neurological or psychiatric diagnoses. A Siemens Verio 3T scanner was used to acquire magnetisation-prepared rapid gradient-echo (MPRAGE) images as follows: TE = 2.98 ms, TR = 2300 ms, TI = 900 ms, flip angle = 9°, field-of-view = 256 mm, 160 slices of 1.0 mm thickness, in-plane resolution = 1.0 x 1.0 mm. The study was approved by the West London NRES Committee and informed, written consent was obtained from each participant before taking part in the research.

2.1.4. **Between-scanner reliability sample**

This dataset comprised 11 participants (male/female = 7/4, mean age at first scan = 30.88 ± 6.16), scanned at two different sites (Imperial College London, Academic Medical Center Amsterdam). The average interval between each scan 68.17 ± 92.23 days, with eight participants being scanned in Amsterdam first, three in London first. High-resolution T1-weighted MRIs were acquired as follows: London Siemens Verio 3T; magnetisation-prepared rapid gradient-echo (MPRAGE), TE = 2.98 ms, TR = 2300 ms, TI = 900ms, flip angle = 9°, field-of-view = 256 mm, 160 slices of 1.0 mm thickness, in-plane resolution = 1.0 x 1.0 mm. Amsterdam Philips Ingenia 3T; sagittal Turbo Field Echo (T1-TFE), TE = 3.1ms, TR = 6.6ms, flip angle = 9°, field-of-view = 270 mm, 170 slices of 1.2 mm thickness, in-plane resolution = 1.1 x 1.1 mm. The study was approved by the West London NRES and the Academic Medical Center Amsterdam institutional review board respectively. Written consent was obtained from each participant before taking part in the research.

## 2.2. Neuroimaging processing

The T1-MRI data for all datasets were processed to generate normalised brain volume maps and 'raw' data appropriate for analysis.

2.2.1. **Normalised brain volume maps**

We followed the protocol as previously outlined (Cole et al., 2015) to generate volumetric maps for use as features in our analysis. Compared to the previous protocol, a minor adaptation was that grey matter (GM) and white matter (WM) images were analysed together, to generate a whole-brain predicted age, as well as age predictions for each tissue. In brief, all images were pre-processed using SPM12 (University College London, London, UK) to segment raw T1 images according to tissue classification (e.g. GM, WM or cerebrospinal fluid) and then generate normalised 3D maps of GM and WM volume, in MNI152 space. Normalisation used DARTEL for non-linear registration and resampling included modulation and 4mm smoothing. This process was applied independently to images from all four datasets, resulting in normalised maps with voxelwise correspondence for all participants.

2.2.2. **Raw data**

While the study aimed to use data in rawest possible form, some minimal pre-processing was carried out to facilitate comparison across different data sources. This included converting from DICOM to Nifti format (using dcm2nii from mricron (Rorden and Brett, 2000)) to be compatible with our in-house software. Raw Nifti files then underwent a rigid registration (i.e. six degrees-of-freedom) to MNI152 space (FMRIB Software Library [FSL] FLIRT, Jenkinson and Smith, 2001), to ensure consistency of orientation (Right, Posterior, Inferior [RPI]). The images were resampled, using cubic spline interpolation, to common voxel sizes and dimensions (1mm$^3$, 182x218x182), as the different contributing studies had acquired data at different dimensions. While not technically in 'raw' form, we assert that this is the rawest form possible for any type of multi-subject analysis, and that the assumptions used here are minimal and unequivocal. Examples of the different types data used in the study are shown in Figure 1.

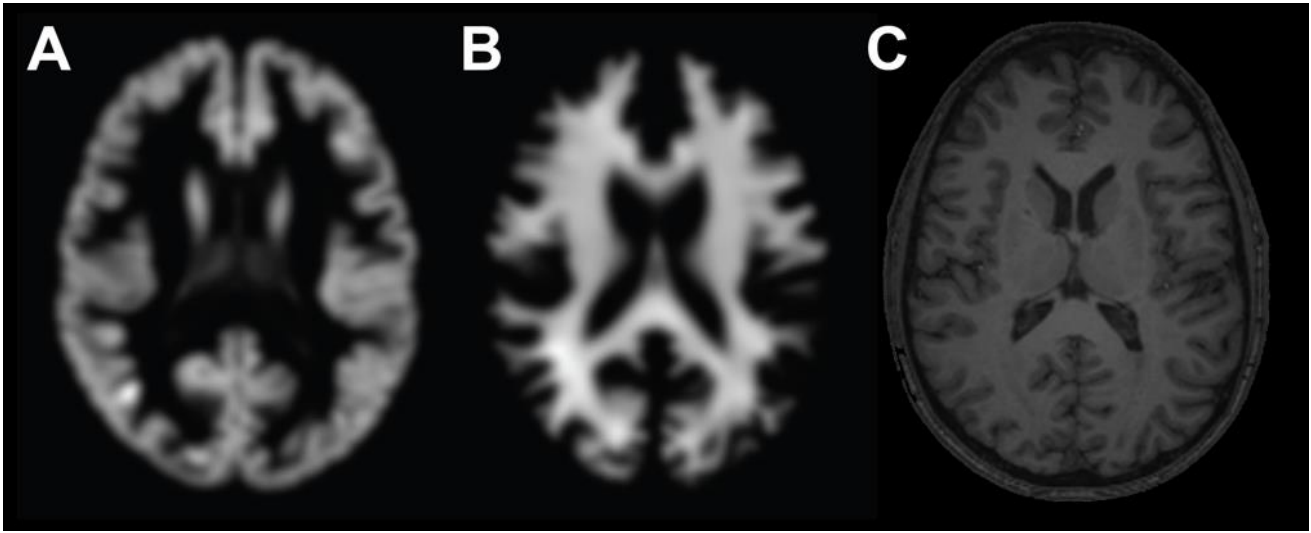

**Fig. 1. Examples of neuroimaging input data for use in age prediction models.** A) Grey matter volumetric map, normalised to MNI152 space using SPM DARTEL for non-linear registration, 4mm smoothing and modulation, in axial view. B) White matter volumetric map, normalised to MNI152 space, in axial view. C) Raw, or minimally-processed, T1-weighted MRI, rigidly-registered to MNI152 space and resampled to a common voxel space.

## 2.3. Machine learning brain age modelling methods

### 2.3.1. Convolutional neural networks

Since their first appearance, CNNs have been very actively investigated, especially in more recent years. Several different network architectures have been proposed, which have enabled to reach state-of-the-art predictive performance in many computer vision and speech recognition tasks . Our hypothesis was that a CNN would provide an appropriate architecture to infer imaging features, from both processed and un-processed brain MRI scans, that optimally predict brain age. When properly trained, CNNs have been shown to be invariant to several variability sources, such as rotation or contrast (Krizhevsky et al., 2012), an aspect that makes them particularly appealing for our application. Given the nature of MR imaging, we have developed a network architecture that uses 3D convolutions , which are appropriate when dealing with fully volumetric images. Recently, 3D convolutional neural networks have been also proposed for Alzheimer's disease classification (Payan and Montana, 2015; Sarraf and Tofighi, 2016), brain lesion segmentation (Kamnitsas et al., 2016) and skull stripping .

Our proposed 3D CNN architecture uses MRI volumes of a size ($z \times h \times w$) as inputs. The specific dimensions, in our applications, are 182 × 218 × 182 when using raw data and 121 × 145 × 121 when using registered GM/ WM data. The output to be predicted is a single scalar representing the biological age. A schematic illustration of the 3D CNN architecture is given in Figure 2. The architecture contains repeated blocks of a (3 × 3 × 3) convolutional layer (with stride of 1), a rectified linear unit (ReLU), a (3 × 3 × 3) convolutional layer (with stride of 1), a 3D batch-normalization layer (Ioffe and Szegedy, 2015), a ReLU and finally a (2 × 2 × 2) max-pooling layer (with stride of 2). The number of feature maps was set to eight in the first block, and was doubled after each max-pooling layer to infer a sufficiently rich representation of the brain. The final age prediction is obtained by using a fully connected layer, which maps the output of the last block to a single output value. For the brain-predicted age using both GM and WM data, we first pre-trained the two individual networks using only GM and WM input data, and then created a single architecture where the highest level blocks of these two networks were joined. A final fully connected layer was then added to predict age using both inputs.

In each application, the network weights were trained by minimizing the MAE using a stochastic gradient descent optimisation algorithm with momentum (Sutskever et al., 2013). Back-propagation was used to compute the gradient of the objective function with respect to all parameters of the model. At the training phase, all datasets were augmented by generating additional artificial training images to prevent model over-fitting. The data augmentation strategy consisted of performing translation (±10 pixels) and rotation (±40 degree), and was found empirically to yield better performance compared to no data augmentation.

All the results reported in Section 2.4 refer to the best out of 3 experiments in which the models were initialised with random parameters and trained end-to-end. The best results were achieved using a learning rate of 0.01 with constant decay of 3% after each epoch, a momentum of 0.9 and weight decay of 0.00005. Training the CNN architectures with only GM or WM input, combined GM and WM input, and raw data input took 18, 42, and 83 hours of training time, respectively, using four GPUs (NVidia TitanX). Importantly, however, testing time in all cases ranged between only 290-940 milliseconds, depending on input type, on a single GPU. All software was written in Torch, a scientific computing framework with support for machine learning algorithms and GPU computing.

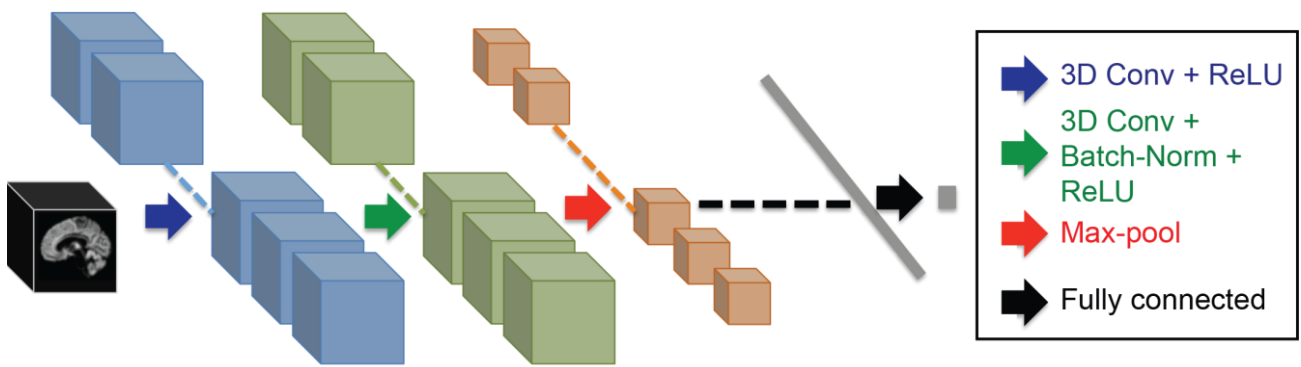

**Fig. 2. Overview of the 3D convolutional neural network architecture.** 3D boxes represent input and feature maps. The arrows represent network operations: blue arrow indicates 3D convolutional operation and a rectified linear unit (ReLU), green arrow indicates 3D convolutional operation, 3D batch normalization and ReLU, red arrow indicates max-pooling operation. Our brain age prediction architecture contains 5 blocks of 3D convolution, ReLU, 3D convolution, 3D batch normalization, ReLU and max-pooling operations and one fully connected layers at the end generate the regression model to output brain-predicted age.

### 2.3.2. Gaussian processes regression

To contextualise the age-prediction performance of CNN, Gaussian Processes Regression (GPR) was used for comparison, as it has previously shown high accuracy in predicting chronological age from T1-MRI data (Cole et al., 2015). A Gaussian Processes (GP) can be thought of a function that extends the multivariate Gaussian distribution, that can be applied over an infinite number of variables. The assumption in GPs is that any finite subset of the data has a multivariate Gaussian distribution. The prior belief about the relationship between variables is informed by definition of these (unlimited number of) multivariate Gaussians in order to generate a model that represents the observed variance. As the multivariate Gaussians can reflect local patterns of covariance between individual points, the combination of multiple Gaussians in a GP can readily model non-linear relationships and is more flexible that conventional parametric models, which rely on fitting global models. GPs can be applied either to categorical data (for GP classification) or continuous data (the GPR approach).

The GPR method was implemented using the Pattern Recognition for Neuroimaging Toolbox (PRoNTo v2.0, www.mlnl.cs.ucl.ac.uk/pronto). Normalised volume images were converted to vectors and the resulting GM and WM vectors concatenated for each subject. A linear kernel representation of these data was then derived by calculating an N-by-N similarity matrix, where each point in the matrix was the dot (scalar) product of two subjects' image vectors. This step retains all the original image variance in a much sparser representation, greatly reducing subsequent computation time. A GPR function was defined, with chronological age as the dependent variable and the image data (in similarity matrix form) as the independent variables, to build a model of healthy structural brain ageing across the adult lifespan. The model was then trained and tested to assess prediction accuracy, using a cross-validation process as outlined below.

## 2.4. Statistical analysis

### 2.4.1. Machine learning age prediction evaluation

Both the CNN and GPR methods were used to predict chronological age using structural neuroimaging as input data. The input data took four different forms; three using normalised brain volume maps (GM only, WM only, GM and WM combined [i.e. concatenated vectors]) and one using raw T1 data. Each learning method was evaluated with each data type, resulting in eight accuracy assessments.

BANC dataset (N = 2001) was used for this stage, and was randomly split into a large training (N = 1601), a validation set (N = 200) and a test set (N = 200). All accuracy assessments reported used predictions made on the test set. Model accuracy was expressed as the correlation between age and predicted age (Pearson's r), total variance explained ($R^2$), MAE and root mean squared error (RMSE).

### 2.4.2. Heritability analysis

Assessment of the heritability of brain-predicted age utilised the TwinsUK sample (N = 62). Using the models trained on the BANC dataset (N = 1601), unbiased age predictions were made for the TwinsUK participants, to generate a brain-predicted age score for each individual. Heritability estimation was performed using Structural Equation Modelling (SEM) . SEM evaluates which combination of additive (A) genetic, common (C) environmental and unique (E) environmental variance components can best explain the observed phenotypic variance and covariance of monozygotic and dizygotic twin data. The importance of individual variance components is assessed by dropping components sequentially from the set of nested models: ACE→AE→E. In choosing between models, variance components are excluded from the selection process if there is no significant deterioration in model fit after the component is dropped, as assessed by the Akaike Information Criterion (Akaike, 1974).

The E component represents random error and must be retained in all models (Rijsdijk and Sham, 2002). Heritability estimates for the AE models are calculated using the formula $h^2 = \frac{a^2}{a^2+e^2}$, where *a* and *e* are the path coefficients of the A and E variance components in the SEM model.

### 2.4.3. Reliability analysis

To calculate the reliability both within- and between-scanner, the intraclass correlation coefficient (ICC) was used. Specifically, this was ICC[2,1] according to Shrout & Fleiss' (1979) notation, to assess absolute agreement between single raters (e.g. scanners). Again using the models trained on the BANC training set (N = 1601), unbiased age predictions were made for each participants' scans in the within-scanner and between-scanner reliability datasets. By subtracting chronological age (at time of scan) from brain-predicted age, a brain-predicted age difference (Brain-PAD) score was calculated. ICC was calculated comparing data from scans approximately four weeks apart (within-scanner sample) and comparing data from a Siemens scanner in London and a Philips scanner in Amsterdam (between-scanner sample).

## 3. Results

### 3.1. Convolutional neural networks accurately predict age using neuroimaging

Analysis showed that our CNN method could accurately predict the chronological age of healthy adults, using either processed volumetric maps or raw T1-MRI data (see Table 1). Prediction accuracy was similar for GPR. The lowest MAE achieved was using GM data and CNN analysis (MAE = 4.16 years), though other predictions were generally comparable. Using single tissues (i.e. GM or WM) did not appreciably alter the prediction accuracy compared to using all available input data for each subject (i.e. GM+WM or raw). The different prediction methods and input data combinations all provided highly accurate estimates. Three example predictions on the test set (N = 200) are shown in Figure 3.

**Table 1. Chronological age prediction accuracy**

| Method | Input data | MAE (years) | r | $R^2$ | RMSE |
|---|---|---|---|---|---|
| **CNN** | GM | 4.16 | 0.96 | 0.92 | 5.31 |
| | WM | 5.14 | 0.94 | 0.88 | 6.54 |
| | GM+WM | 4.34 | 0.96 | 0.91 | 5.67 |
| | Raw | 4.65 | 0.94 | 0.88 | 6.46 |
| **GPR** | GM | 4.66 | 0.95 | 0.89 | 6.01 |
| | WM | 5.88 | 0.92 | 0.84 | 7.25 |
| | GM+WM | 4.41 | 0.96 | 0.91 | 5.43 |
| | Raw | 11.81 | 0.57 | 0.32 | 15.10 |

**MAE = mean absolute error, r = Pearson's r from correlation between chronological age and brain-predicted age, RMSE = root mean squared error, GM = Grey Matter, WM = White Matter.**

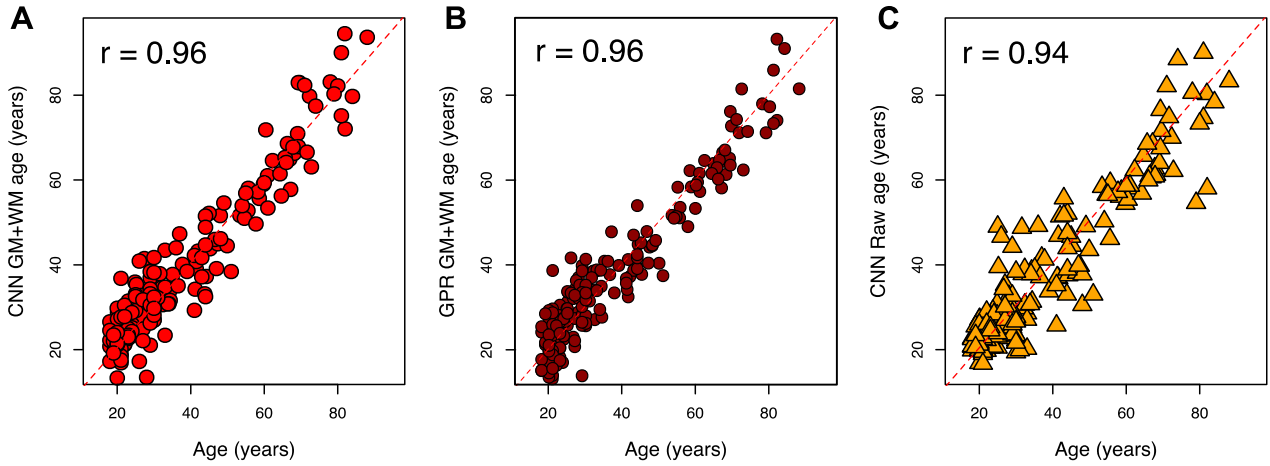

**Fig 3. Accuracy of CNN and Gaussian Processes Regression for age prediction.** Scatterplots depict chronological age (x-axis) and brain-predicted age (y-axis) on the test-set subjects from the BANC dataset (N = 200). A) Brain-predicted ages derived used GM maps as input data for the CNN method. B) Brain-predicted ages derived using GM maps as input data for the Gaussian Processes Regression (GPR) method. C) Brain-predicted ages derived using raw T1-MRI as input for the CNN method. R-values in all plots are the Pearson's correlation coefficient of brain-predicted age with chronological age.

### 3.2. Brain-predicted age is significantly heritable

Brain-age heritability estimates were consistently high, irrespective of the input data and predictive model employed (Table 2). The highest estimate was produced for the CNN-predicted ages using GM+WM data. The only exception was the heritability of the raw-based predictions, which was considerably lower, compared to the other values. Notably, the heritability of brain-predicted age was reduced after controlling for the effect of chronological age. The same combination of model and input data again provided the highest estimate of 0.66. For all prediction methods, the AE models proved to have the best fit (i.e. lowest AIC).

**Table 2. Heritability estimates from the AE SEM models for different brain-predicted age methods**

| Method | GM | WM | GM+WM | Raw |
| --- | --- | --- | --- | --- |
| *Unadjusted* | | | | |
| **CNN** | 0.74 ± 0.09 | 0.78 ± 0.07 | 0.84 ± 0.05 | 0.62 ± 0.10 |
| **GPR** | 0.78 ± 0.07 | 0.81 ± 0.06 | 0.82 ± 0.06 | 0.64 ± 0.10 |
| *With age-correction* | | | | |
| **CNN** | 0.55 ± 0.11 | 0.65 ± 0.10 | 0.66 ± 0.09 | 0.50 ± 0.12 |
| **GPR** | 0.55 ± 0.11 | 0.60 ± 0.10 | 0.58 ± 0.11 | 0.64 ± 0.10 |

Heritability estimates are given by $h^2 = \frac{a^2}{a^2+e^2}$, where *a* and *e* are the path coefficients of the A and E variance components in the SEM model, ± the standard errors of the estimates. GM = grey matter, WM = white matter, CNN = convolutional neural network, GPR = Gaussian processes regression.

### 3.3. Brain-predicted age difference is highly reliable

Brain-PAD scores were generally highly reproducible, whether using CNN or GPR to generate brain-predicted ages. This was the case for both within-scanner (i.e. scanner test-retest) reliability and between-scanner (i.e. multi-site) reliability (see Table 3, Figures 4 and 5). All the different combinations of input data (GM, WM, GM+WM, raw) and prediction method (CNN, GPR) resulted in a significant ICC for reliability, with the one exception of WM using CNN. Broadly speaking, the within-scanner reliability estimates were higher than the between-scanner estimates, and this difference was more pronounced for CNN brain-PAD than GPR brain-PAD. For the latter, between-scanner reliability for GM and GM+WM was as high as within-scanner reliability. Notably, the within-scanner reliability for raw data using CNN was very high (ICC = 0.94), though substantially reduced when comparing estimates from two scanners (ICC = 0.66).

**Table 3. Within-scanner and between-scanner reliability estimates of brain-predicted age difference**

| Method | Dataset | GM | WM | GM+WM | Raw |
|---|---|---|---|---|---|
| **CNN** | Within | 0.90 [0.76, 0.96] | 0.97 [0.90, 0.99] | 0.90 [0.77, 0.96] | 0.94 [0.86, 0.98] |
|  | Between | 0.83 [0.49, 0.95] | 0.51 [-0.08, 0.84] | 0.85 [0.55, 0.96] | 0.66 [0.17, 0.89] |
| **GPR** | Within | 0.96 [0.90, 0.98] | 0.98 [0.94, 0.99] | 0.97 [0.92, 0.99] | 0.99 [0.97, 0.99] |
|  | Between | 0.96 [0.88, 0.99] | 0.77 [0.12, 0.94] | 0.92 [0.74, 0.98] | 0.56 [-0.02, 0.86] |

**All figures in the table are intraclass correlation coefficients (ICC) and 95% confidence intervals. GM = grey matter, WM = white matter, CNN = convolutional neural network, GPR = Gaussian processes regression.**

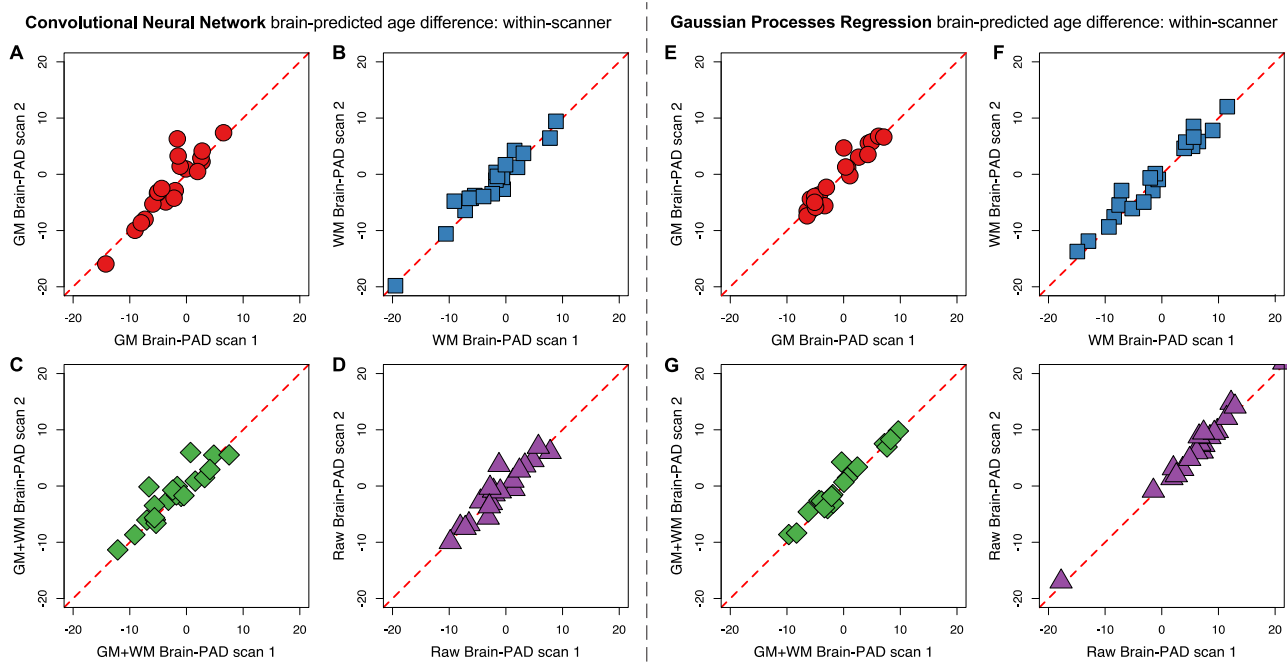

**Fig. 4. Within-scanner reliability for Convolutional Neural Networks and Gaussian Processes Regression.** Figure shows the correspondence between brain-predicted age difference (Brain-PAD) based on scans acquired four weeks apart on the same scanner (Siemens Verio 3T) on N = 20 individuals, with scan 1 on the x-axis and scan 2 (after four weeks) on the y-axis for all plots. A) Brain-PAD score based on GM maps using CNN. B) Brain-PAD score based on WM maps using CNN. C) Brain-PAD scored based on GM and WM maps combined using CNN. D) Brain-PAD scored based on raw T1-MRI using CNN. E) Brain-PAD score based on GM maps using Gaussian Processes Regression (GPR). F) Brain-

PAD score based on WM maps using GPR. G) Brain-PAD score based on GM and WM maps combined using GPR. The red dashed line in all plots is the line of identity.

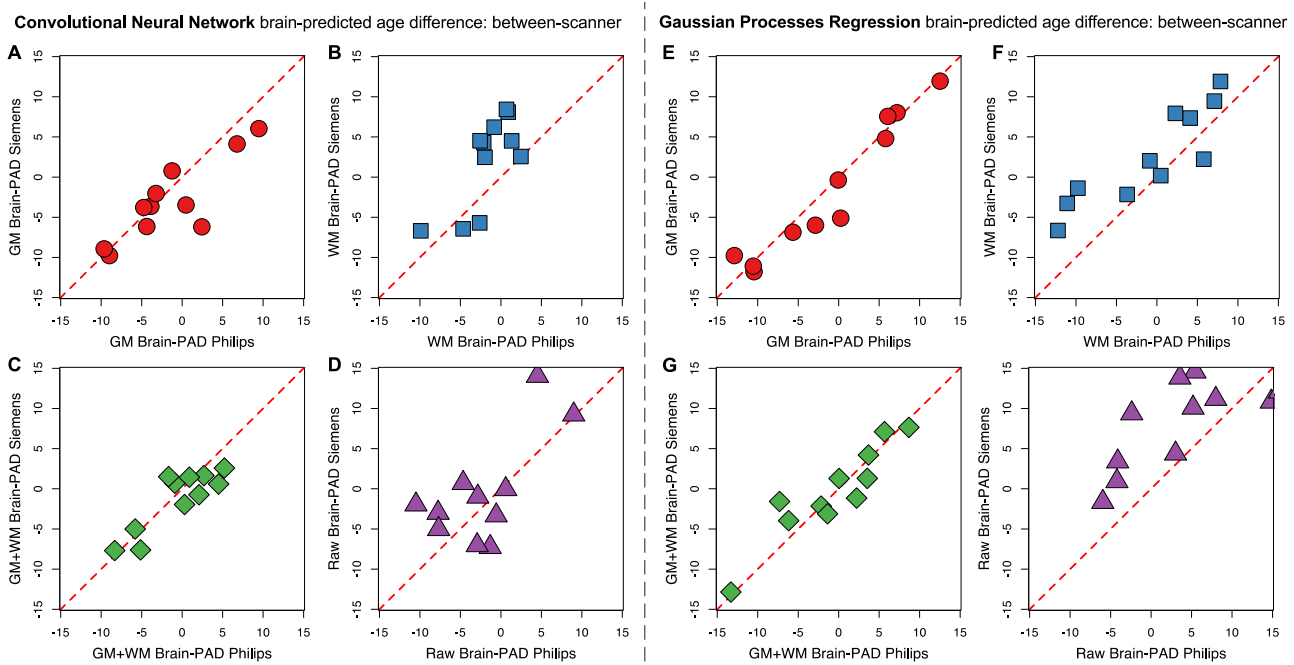

**Fig. 5. Between-scanner reliability for Convolutional Neural Networks and Gaussian Processes Regression.** Figure shows the correspondence between brain-predicted age difference (Brain-PAD) scores based on scans acquired on two different scanner systems (Siemens Verio 3T and Philips Intera 3T) in N = 11 individuals, with the Philips scan on the x-axis and Siemens scan on the y-axis for all plots. A) Brain-PAD score based on GM maps using CNN. B) Brain-PAD score based on WM maps using CNN. C) Brain-PAD scored based on GM and WM maps combined using CNN. D) Brain-PAD scored based on raw T1-MRI using CNN. E) Brain-PAD score based on GM maps using Gaussian Processes Regression (GPR). F) Brain-PAD score based on WM maps using GPR. G) Brain-PAD score based on GM and WM maps combined using GPR. The red dashed line in all plots is the line of identity.

## 4. Discussion

Using 3D convolutional neural networks, we accurately estimated chronological age from raw T1-weighted MRI brain scans of healthy adults. The accuracy of CNN for age prediction was also high when using processed GM and WM voxelwise images, and was comparable with age estimations made using GPR. Brain-predicted age estimates were significantly heritable and showed high levels of within-scanner and between-scanner reliability. These findings support the idea that deep learning methods can generate a viable biomarker of brain ageing: brain-predicted age.

This study is the first illustration that 3D convolutional neural networks can be used to accurately estimate chronological age from neuroimaging data. CNN-based predictions were equivocal compared to a previously employed method, GPR (Cole et al., 2015), and the predictions highly correlated. When using voxelwise images representing GM and WM volume, both methods were able to predicted age with < 5 years MAE. Importantly, CNN-based age prediction accuracy was equally high when using raw (or minimally pre-processed) T1 images as input. This means that sufficient age-related features can be extracted from a 3D image to enable accurate out-of-sample prediction and obviates the necessity of image pre-processing. This brings two key benefits, specifically: 1) removal of additional assumptions that are required to pre-process image data; 2) increasing the feasibility that such an approach could be used in real-time, or near real-time, to augment clinical decision making.

Data pre-processing is almost ubiquitous in neuroimaging, including in previous studies using brain-based age predictions . Multiple different options are available from different software packages for each stage of pre-processing, including bias-field correction, removal of non-brain tissue, tissue classification, motion correction, artefact removal, linear registration, non-linear registration, target image (e.g., atlas, average template), interpolation method and smoothing kernel. While we opted to using SPM here, the relative merits of different approaches are keenly debated . In the absence of a consensus, the ability to model outcome variables without conducting any of these steps is attractive. As the choice of pre-processing pipeline will undoubtedly influence any derived measures, using raw data for prediction removes a key source of variance. Moreover, as the assumptions underlying many of the different pre-processing steps are often not met when dealing with clinical populations containing individuals with atypical brain morphology, using raw data also removes additional confounds and potential biases.

A key goal of neuroimaging research is to make tools for clinical application, that can provide objective and reliable information that clinicians can use to help when treating brain diseases. One element of this goal is in producing real-time methods that generate interpretable outputs from imaging data for immediate use in clinical decision making. Image pre-processing can take >24 hours, hence removing this step represents a substantial acceleration of the pipeline necessary to deliver information to the clinician. Admittedly, the training phase of a deep artificial neural network is computationally-intensive and time-consuming. However, once trained, the model can be applied to new data in a matter of seconds. Given the right software implementation, brain-predicted age data could be made available to a clinician while the patient is still in the scanner. In our study, minimal processing was used when training/testing the CNN algorithm, only to ensure consistent image orientation and voxel dimensions between images. These processes require very limited assumptions and could readily be automated into MR scanner software.

CNN brain-predicted age was significantly heritable, as were all prediction methods, indicating that genetic relatedness influences brain-predicted age. This is important as it provides a degree of external validity. If brain-predicted age were merely a reflection of disease-related atrophy or driven by noise, then the additive genetic models would not significantly account for the observed data. This supports further use of brain-predicted age as a biomarker of brain ageing. Moreover, as previous reports have indicated that brain-predicted age relates to measures of cognitive performance , it could potentially be used to predict risk of future cognitive decline and risk of subsequent dementia. That our measure of brain ageing is under some genetic control corroborates research indicating that cognitive ageing is also influenced by genetic factors . Intuitively it follows that brain ageing (i.e., underlying anatomical changes) and cognitive ageing (i.e., manifest behavioural changes) must be linked. Therefore, our findings, along with previous demonstrations of the heritability of brain structure , motivate research into specific genes which may influence rates of brain and cognitive ageing. Such genes have the potential to offer novel targets for pharmacological interventions aimed at reducing the risk of age-associated neurodegeneration and cognitive decline, as well as even slowing the rate of brain ageing itself.

We also observed that the heritability of brain-predicted age decreased when chronological age was taken into account. This age dependency indicates that genetic influences on brain-predicted age are reduced with increasing age, with a corresponding increase in unique environmental effects. This is in line with previous research into the heritability of brain volumes and has also been demonstrated with regard to cognitive function (Lee et al., 2010). Potentially, the cumulative exposure to environmental factors over a lifetime eventually outweigh the inherited genetic factors that influence brain structure and cognitive function. This certainly has implications for genetic studies of brain ageing, including how studies are designed and samples are selected, however, our limited number of twins in the current study does not permit an exhaustive analysis of how heritability changes with age.

Brain-predicted age was highly reproducible. Reliability estimates varied for different combinations of input data and algorithm, however within-scanner test-retest reliability was high (ICC ≥ 0.90) for all analysis, even using raw data. This is crucial for any measure to be used in longitudinal studies, and has an important bearing on the sample sizes required to detect significant effects on repeated measures (Guo et al., 2013). The test-retest reliability of T1-MRI derived brain structural measures has been demonstrated , and our results are consistent with these estimates. This high reproducibility supports the use of brain-predicted age in longitudinal research or potentially clinical settings.

Regarding between-scanner reliability for brain-predicted age, the results were more contrasting. GPR between-scanner reliability was generally higher than for CNN, and GM was generally better than WM. This agrees broadly with previous studies investigating the reliability of T1-MRI measures in multi-centre settings . Between-scanner reliability was substantially reduced for raw data. Potential explanations for this include differences in contrast-to-noise ratio observed between different vendors on T1-MRI, or differences in shimming effectiveness between different scanners. The pre-processing steps used to generate normalised GM and WM images largely remove the effects of inconsistent gradient distortions, by carrying out bias-field correction and estimating tissue probabilities. Conversely, the CNN architecture may be characterising these are explanatory features at a given level of the model. Currently, it would seem that deep learning models using raw data are most appropriate for longitudinal studies of brain ageing on the same scanner, where the issue of image heterogeneity due to inter-scanner variability is not present. Therefore, there are likely still benefits to data pre-processing when pooling data from multiple scanners, as would be the case in a multi-centre study, to remove clearly identifiable sources of technical variability that are unrelated to brain ageing.

There are some limitations of the study to consider. The sample size for the heritability estimates was small, particularly regarding the numbers of dizygotic twins included and the sample was composed of females, hence we cannot readily extrapolate to males. However, the confidence intervals take sample size into account and while the precision behind the estimates could be greatly improved by increasing numbers, the evidence that at least some of the variance in brain-predicted age is moderated by genetic relatedness remains valid. Another limitation is that our between-scanner reliability analysis only used data from two scanners, with the same field strength. To build up a comprehensive picture of the influence of scanner system on brain-predicted age, further varieties of scanner should be analysed.

## 5. Conclusions

Deep learning models based on T1-MRI can accurately predict chronological age in healthy individuals. This can be achieved using raw MRI data, with a minimum of processing necessary to generate an accurate age prediction. These estimates of brain-predicted age are also significantly heritable, giving external, genetic, validity to the measure and motivating its use in genetic studies of brain ageing. Finally, our analysis showed the brain-predicted age is highly reliable and thus appropriate for use in both longitudinal and multi-centre studies. Brain-predicted age has the potential to be used as a biomarker to investigate the brain ageing process and how this relates to cognitive ageing, neurodegeneration and age-associated brain diseases.

## Acknowledgements

TwinsUK is funded by the Wellcome Trust, Medical Research Council, European Union, the National Institute for Health Research (NIHR)-funded BioResource, Clinical Research Facility and Biomedical Research Centre based at Guy's and St Thomas' NHS Foundation Trust in partnership with King's College London.